\documentclass[10pt,twocolumn]{article} 
\usepackage{simpleConference}
\usepackage{times}
\usepackage{graphicx}
\usepackage{amssymb}
\usepackage{url,hyperref}
\usepackage[numbers]{natbib}

\usepackage{enumerate}
\usepackage{amsmath} 
\usepackage{amssymb}
\usepackage{subcaption}
\usepackage{caption} 
\usepackage{hyperref}
\usepackage{microtype}

\usepackage{booktabs} 
\usepackage{xcolor}

\usepackage[switch]{lineno}

\definecolor{orange}{rgb}{1, 0.5, 0.0}

\newcommand{\NCA}{f_{NCA}} 
\newcommand{\neigh}{N}  
\newcommand{\ltwo}{\mathcal{L}_{2}}

\begin{document}

\title{Generative Adversarial Neural Cellular Automata}

\author{Maximilian Otte ~~~ Quentin Delfosse ~~~ Johannes Czech ~~~ Kristian Kersting\\
\\
AIML Lab, Technical University Darmstadt \\
Darmstadt, Germany \\
\today
\\
\\
MaximilianFinn.Otte@stud.tu-darmstadt.de  \\
}

\maketitle
\thispagestyle{empty}

\begin{abstract}
\noindent Motivated by the interaction between cells, the recently introduced concept of Neural Cellular Automata shows promising results in a variety of tasks.
So far, this concept was mostly used to generate images for a single scenario.
As each scenario requires a new model, this type of generation seems contradictory to the adaptability of cells in nature.
To address this contradiction, we introduce a concept using different initial environments as input while using a single Neural Cellular Automata to produce several outputs.
Additionally, 
we introduce GANCA, a novel algorithm that combines Neural Cellular Automata with Generative Adversarial Networks, allowing for more generalization through adversarial training.

The experiments show that a single model is capable of learning several images when presented with different inputs, and that the adversarially trained model improves drastically on out-of-distribution data compared to a supervised trained model.
\end{abstract}

\section{Introduction}
\label{Introduction}
Regeneration of body parts is an example of the fundamental self-organization of cells. Even though each cell can only interact with its immediate surroundings, it acts accordingly to its position inside the body. Cells are able to react and re-organize depending on external stimulation or environmental changes \cite{karsentiSelforganizationCellBiology2008}.
These biological phenomena are one of the motivations for Neural Cellular Automata (NCAs).

Shortly after the paper introducing NCAs by 
\citet{mordvintsevGrowingNeuralCellular2020}, several papers were published showcasing the performance of this architecture on different tasks.
Growing and robustly repairing images \cite{mordvintsevGrowingNeuralCellular2020}, texture generation \cite{niklassonSelfOrganisingTextures2021}, classification \cite{randazzoSelfclassifyingMNISTDigits2020} or pixel wise segmentation \cite{sandlerImageSegmentationCellular2020}, show the high diversity of applications for Neural Cellular Automata.

In all previous experiments, NCAs are trained to perform well in a single situation, producing good-looking results, but defying the nature of the biological motivation for the model. 
This is why this paper starts of by illustrating that the NCA is 
capable of generating different images depending on the initial image.
These generalization capabilities will then be further tested on unseen initial images through validation data and out-of-distribution data.
Furthermore, we introduce a new type of NCA, the Generative Adversarial Neural Cellular Automaton~(GANCA), combining the adversarial training of a GAN structure with the generative capabilities of an NCA, to increase the performance on out-of-distribution data.

\section{Foundation}
\begin{figure}
    \centering
    \includegraphics[width=0.95\linewidth]{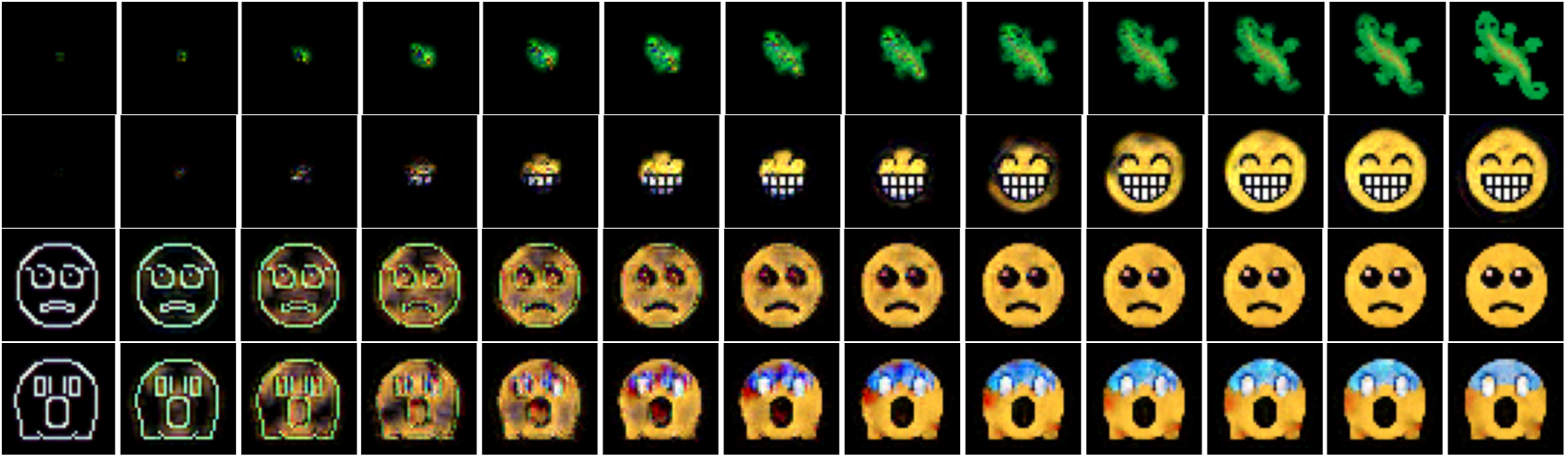}
    \caption{Frame by frame output of three different NCAs growing emojis. Each image shows progress of frames, starting from a blank (as in \citet{mordvintsevGrowingNeuralCellular2020}) or edge image (ours) from the training set. The last two emojis were generated from the same NCA.
    }
    \label{fig:single_emoji_frames}
\end{figure}

A Neural Cellular Automaton, spreads and computes local information while using the same program for each local part.
An example of an NCA generating images frame by frame can be seen in Figure \ref{fig:single_emoji_frames}.

\begin{figure*}[t]
\centering
  \centering
  \includegraphics[width=.85\linewidth]{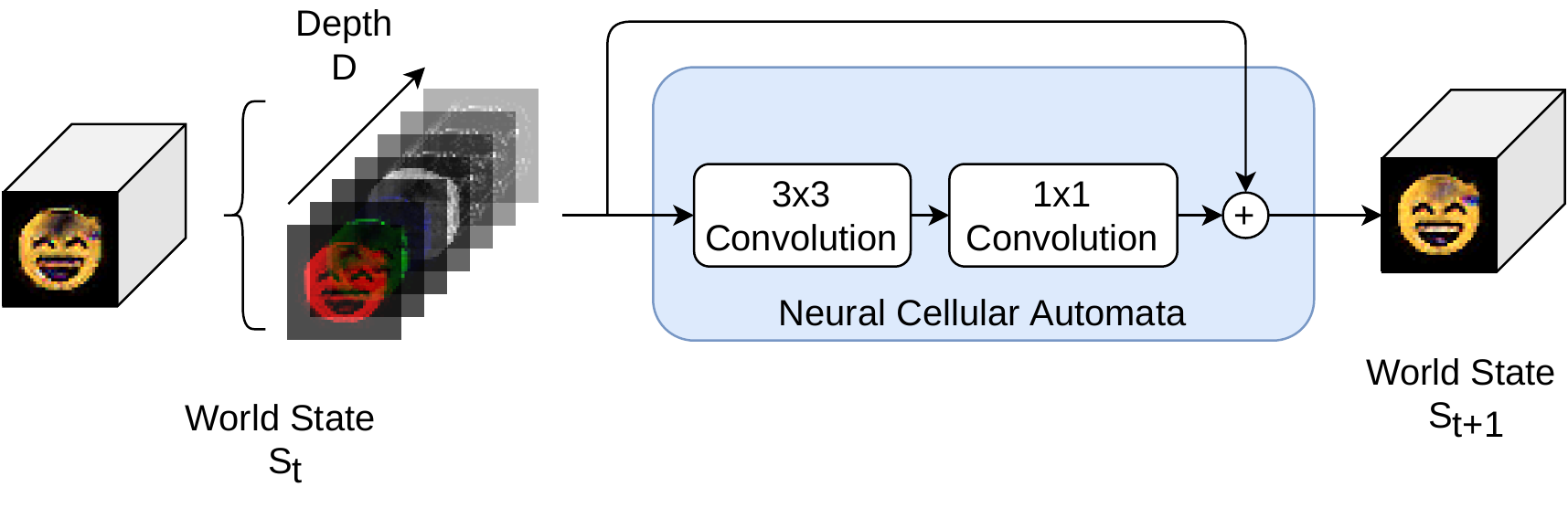}
  \caption{Sketch of the implementation of a single iteration of a simple Neural Cellular Automaton.
  The world state contains 4 channels of R.G.B.A. information, while additional channels contain latent state information. We use a world state depth of 16, resulting in around 20k parameters for the NCA. Increasing the depth drastically increases the number of parameters.
  }
  \label{fig:simple_nca}
\end{figure*}

Intuitively speaking, an NCA follows the same mandatory restrictions:
\newpage
\begin{enumerate}[(i)]
    \item \label{nca:i} Every \textbf{cell} only sees its local neighborhood.
    \item \label{nca:ii} Every \textbf{cell} runs the same \textbf{program}.
\end{enumerate}
To propagate the local information globally, the NCA spreads this information throughout time.
In most cases the \textbf{world} consists of a grid of \textbf{cells}. 
An exemplary grid world is shown in Figure~\ref{fig:single_emoji_frames}, where each pixel represents one cell.

A \textbf{cell} has a state vector containing information of this cell, and can only observe its local neighborhood by seeing the state vectors of the closest cells.
Using the \textbf{program}, every cell will use this information to update their state vector. This way, information can be propagated through the whole world.

All NCA architectures inherently follow these statements. Some architectures add additional constraints, while others only differ in the program that each cell will execute. 

\subsection{Formalization}

Using the restrictions, we propose a new formalization for the NCA architecture.
Assuming a grid world\footnote{A grid world is not mandatory. However, assuming a grid world simplifies the formalization, making it easier to follow.},
$S_{t} \in \mathbb{R}^{W \times H \times D}$ defines the world state at time step $t$, where $W$ is the width, $H$ the height and $D$ the depth of the world. The depth~$D$ also describes the size of the state vectors for each cell.
The state vector of a cell at position $(x,y) \in W\times H$, during time step $t$ will be written as $\mathbf{c}_{xy}^{t} \in \mathbb{R}^{D}$.

This will lead to a definition of the world state $S_{t}$:
\begin{align*}
    S_{t} := 
    \begin{pmatrix}
    \mathbf{c}_{0,0}^{t} & \dots & \mathbf{c}_{0,W}^{t} \\
    \vdots & \ddots & \vdots \\
    \mathbf{c}_{H,0}^{t} & \dots & \mathbf{c}_{H,W}^{t}
    \end{pmatrix}.
\end{align*}
Using these notations, the first statement (\ref{nca:i}) can be written as:
\begin{align*}
\begin{split}
    \neigh(\mathbf{c}_{x,y}^{t}) &:= 
    \begin{pmatrix}
    \mathbf{c}_{x-1, y-1}^{t} & \mathbf{c}_{x, y-1}^{t} & \mathbf{c}_{x+1, y-1}^{t} \\
    \mathbf{c}_{x-1, y}^{t} & \mathbf{c}_{x, y}^{t} & \mathbf{c}_{x+1, y}^{t} \\
    \mathbf{c}_{x-1, y+1}^{t} & \mathbf{c}_{x, y+1}^{t} & \mathbf{c}_{x+1, y+1}^{t}
    \end{pmatrix}, \\
    \mathbf{c}_{x,y}^{t+1} &:= g(
    \mathbf{c}_{x,y}^{t},
    \neigh(\mathbf{c}_{x,y}^{t})),
\end{split}
& (\ref{nca:i})
\end{align*}
where $\neigh$ is a neighborhood function returning only the local information around a cell, the $3 \times 3$ neighborhood in this case.
The program, or local transition function, is denoted here as $g$.
This can be used to define the transition function of the NCA as:
\begin{displaymath}
\NCA(\mathbf{c}_{x,y}^{t}) := g(
    \mathbf{c}_{x,y}^{t},
    \neigh(\mathbf{c}_{x,y}^{t})).
\end{displaymath}
The second statement (\ref{nca:ii}) can now be defined through the update rule of the whole world state:
\begin{align*}
\begin{split}
    \NCA(S_{t}) &:= 
    \begin{pmatrix}
    \NCA(\mathbf{c}_{0,0}^{t}) & \dots & \NCA(\mathbf{c}_{0,W}^{t}) \\
    \vdots & \ddots & \vdots \\
    \NCA(\mathbf{c}_{H,0}^{t}) & \dots & \NCA(\mathbf{c}_{H,W}^{t})
    \end{pmatrix},\\
    S_{t+1} &:= \NCA(S_{t}).
\end{split}
& (\ref{nca:ii})
\label{math:transition}
\end{align*}
Using zero-padding, the values outside the boundaries, e.g. $\mathbf{c}_{-1,-1}^{t}$, are set to 0. Throughout the matrix, the same function $\NCA$ is applied to every element, running the same program $g$ across all cells.

Using these definitions, one of the final images in Figure~\ref{fig:single_emoji_frames} can be written as:
\begin{align*}
    S_{60} = \underbrace{[\NCA \circ \dots \circ \NCA]}_\text{60 times}(S_{0}) \ .
\end{align*}

The initial world state $S_{0}$ is, for example, the blank image with only modified pixels in the middle of the image.
The same $\NCA$ transition will be applied 60 times to result in $S_{60}$.

\subsection{Implementation}
The NCA can be implemented through existing deep learning structures, as it simply is a ResNet block \cite{heDeepResidualLearning2016} with two convolutional layers.
The first layer is the perception layer propagating the information of a cell to the neighboring cells through $3\times3$ convolution, while the following layer(s) are $1\times1$ convolutions for additional computation. 
A single \textbf{iteration} is a single pass through this block, which is visualized in Figure \ref{fig:simple_nca}.

It is possible to train the NCA in a supervised fashion through a pixel-wise L2 loss, e.g. comparing the first 4 layers produced by the
NCA, representing the RGBA channels, to the target image. 
Additionally, in order to keep the image stable over a longer period of time, \citet{mordvintsevGrowingNeuralCellular2020} introduced \textbf{persistence}.
The goal of persistence is the same as finding an update rule such that for every $i \in \mathbb{N}$ the following equation holds
\begin{equation*}
\ltwo (S_{t_{n}+i}) = \ltwo (S_{t_{n}}) + \epsilon \; ,
\end{equation*}
where $\ltwo$ is an exemplary loss function, calculating the loss to the target image at time step~$t_{{n}+i}$, with $\epsilon \approx 0$.

Persistence can be accomplished, for example, by having a chance of using the output of the final NCA iteration of the previous training step as input for the first iteration of the next training step.
This way, the NCA additionally has to keep the image stable over a possibly infinite period of time. 

An exemplary implementation of an NCA and all the experiments presented in this paper can be found on GitHub\footnote{Link to be announced}.

\section{Related Work}
Beginning with a collection of articles about differentiable self-organizing systems\footnote{\url{https://distill.pub/2020/selforg/}}, \citet{mordvintsevThreadDifferentiableSelforganizing2020} motivate a new approach to self-organization in systems.

In the first article of this thread, \citet{mordvintsevGrowingNeuralCellular2020} introduce the concept of Neural Cellular Automata, where
they grew emojis from a blank seed using a single NCA. 
The objective was to start from a blank image and start to grow the emoji step by step. This was further extended by regrowing the image after it has been damaged by a user interaction.

Their follow-up article uses the NCA for the MNIST classification problem \cite{randazzoSelfclassifyingMNISTDigits2020}.
This is done by using the information of the state vector from each cell, such that each cell will classify itself as one of the classes from the MNIST set. 
Even though the task is vastly different to the first NCA application, the training is very similar. 

The third paper in the timeline, by \citet{sandlerImageSegmentationCellular2020}, is not part of the self-organizing thread.
In this paper, the task is to use the NCA for pixel-wise segmentation and classification.
With this paper, they introduced several new ways of training the NCA, which will be partly used throughout the experiments.

The most recent article by \citet{niklassonSelfOrganisingTextures2021} focused on creating textures with NCA architectures.
Here, they used the information of a VGG-16 (\cite{vggnet}) network to apply a loss on the generated texture, allowing for the NCA to generate good-looking textures.

\section{Methods}
This section begins by extending the idea of generating images with an NCA, by producing multiple target images with a single NCA. 
Moreover, this training concept will then be improved in generalization by adversarially training the NCA, creating the Generative Adversarial Neural Cellular Automata (GANCA).

\subsection{Multiple target images}
\label{sec:mult_target}

\begin{figure}
\centering
  \centering
  \includegraphics[width=.85\linewidth]{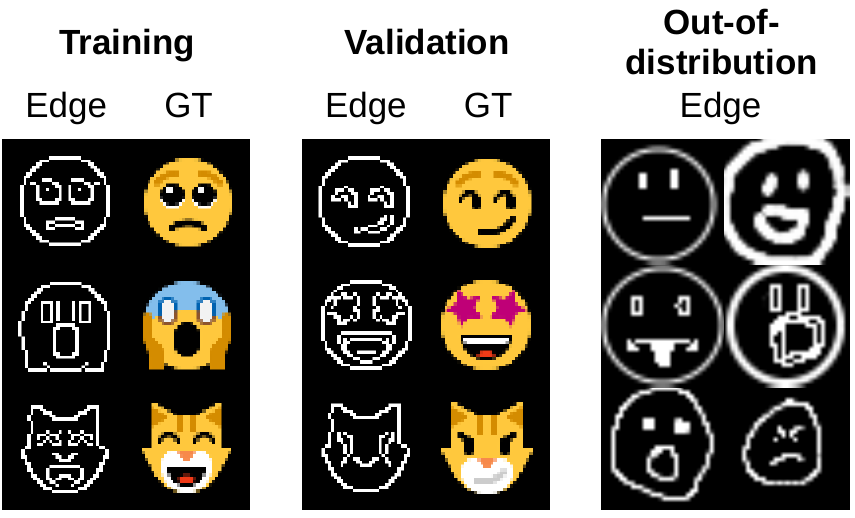}
  \caption{Example emojis for training, validation and out-of-distribution data. ``Edge'' is the input image with only edge information, while ``GT'' shows the ground truth image.
  Every ground truth image is an emoji from the Windows 10 Segoe UI Emoji font.
  The out-of-distribution data are hand drawn images, resembling faces, but vastly different to the training and validation images.}
  \label{fig:validation_data_microsoft}
\end{figure}

\begin{figure*}
\centering
  \centering
  \captionsetup{width=.95\linewidth}
  \includegraphics[width=.85\linewidth]{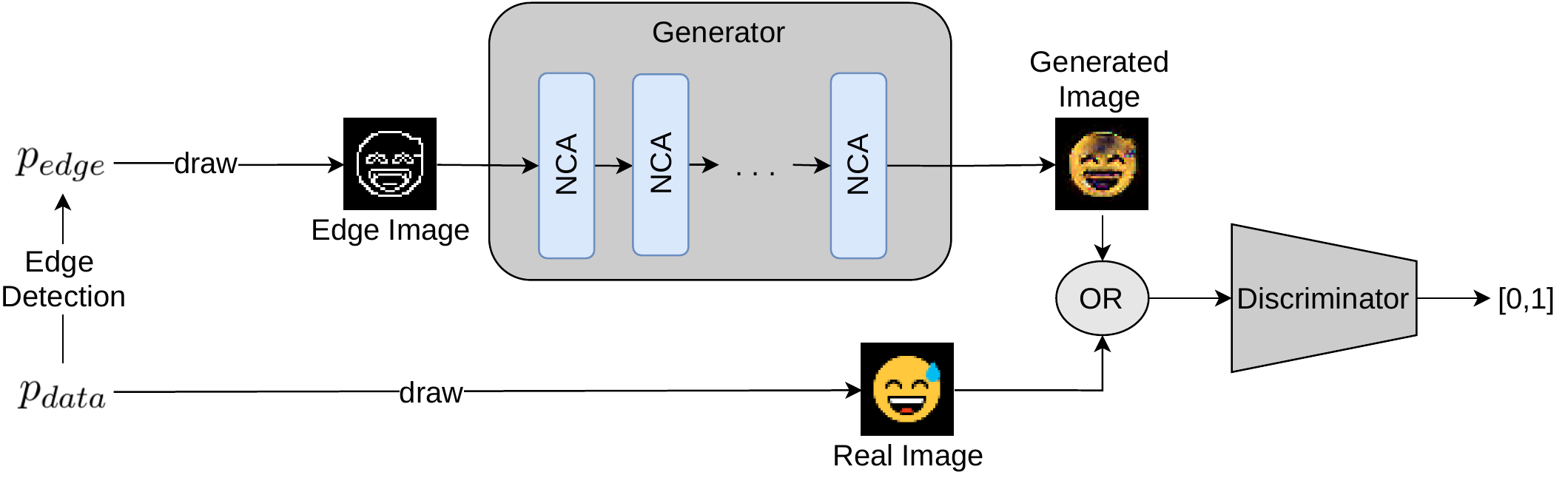}
  \caption{Visualization of the Generative Adversarial Neural Cellular Automata (GANCA) architecture with a very similar setup compared to a standard GAN architecture.
  The generator now uses several iterations of the same NCA to generate an image. Additionally, the input for the generator is an image, which is in our task an edge image generated from the real images.
  }
  \label{fig:ganca}
\end{figure*}

To generate different images based on the input, the input has to contain information of the target image. 
Our approach here is to change the dataset for training to contain a ground truth image as the target and an image consisting of the edges of the ground truth image, which will be used as the input.

The edge image is obtained by using an edge detection algorithm, a 
Canny filter~\cite{canny1986computational} in this case, on the ground truth image.
Examples are shown in Figure~\ref{fig:validation_data_microsoft}.

This way, the NCA has a chance to differentiate between the shape of the input image to generate different fully colored emojis. 
Furthermore, the performance will be tested on a validation set from the same distribution and on an out-of-distribution dataset.

The out-of-distribution images are hand drawn edges which are not perfectly round or have the same line width etc. 
This is used to test how the NCA will react on images vastly different from the training examples.

\subsection{Generative Adversarial Training}
\label{sec:ganca}
The generalization performance on out-of-distribution inputs, after training on multiple target images,
can be further improved through partly\footnote{As the real images are used to generate the initial image for the NCA, the task is not fully unsupervised. This step is not mandatory, and edge images not directly related could be used as well.}
unsupervised training.
To train unsupervised, we use adversarial training from GAN architectures \cite{goodfellowGenerativeAdversarialNets}, where several NCA steps replace the generator. 
Because the NCA operates on images, the input also changes to an image, instead of a random value vector.
In our case, the GAN will be trained on multiple emoji faces, as used in section~\ref{sec:mult_target}.
As the goal of the training is to transform an edge image into a colored version, other GAN architectures show strong similarities.
Because the edge image is based on a specific emoji, the GAN structure can be considered conditional~\cite{mirzaConditionalGenerativeAdversarial2014}.
However, the GANCA also shows strong relations to the Image2Image GAN~\cite{pix2pix2017}, or Cycle GANs~\cite{CycleGAN2017} in general.

The  Generative Adversarial Neural Cellular Automata (GANCA) architecture concept is explained in Figure \ref{fig:ganca}. 
This is very similar to a standard GAN structure, with the main difference being the generator and its input. 
The generator in the GANCA uses an NCA to update the input edge image step by step with only local information to produce an image. 
The input edge image is based on the set of real images provided for the discriminator reduced to edges.
The NCA uses a random amount of iterations in a defined range, e.g. between 50 and 60, to produce the image. A single iteration is a single pass through an NCA block, as visualized in Figure \ref{fig:simple_nca}.

\begin{figure*}
\begin{subfigure}{.51\textwidth}
    \centering
    \includegraphics[width=.99\linewidth]{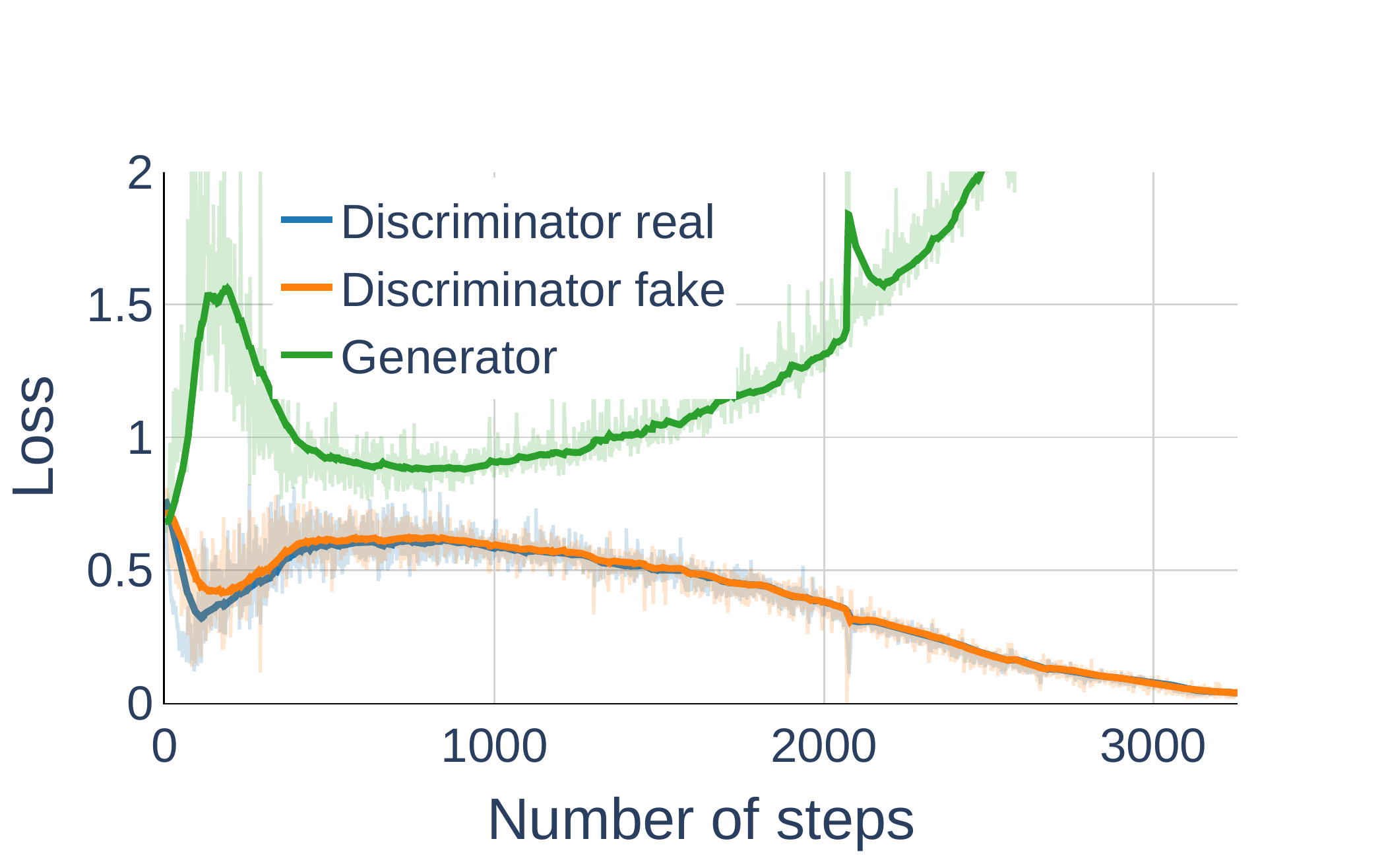}
    \centering
    \caption{Simple GANCA without any modifications}
\end{subfigure}
\begin{subfigure}{.51\textwidth}
    \centering
    \includegraphics[width=.99\linewidth]{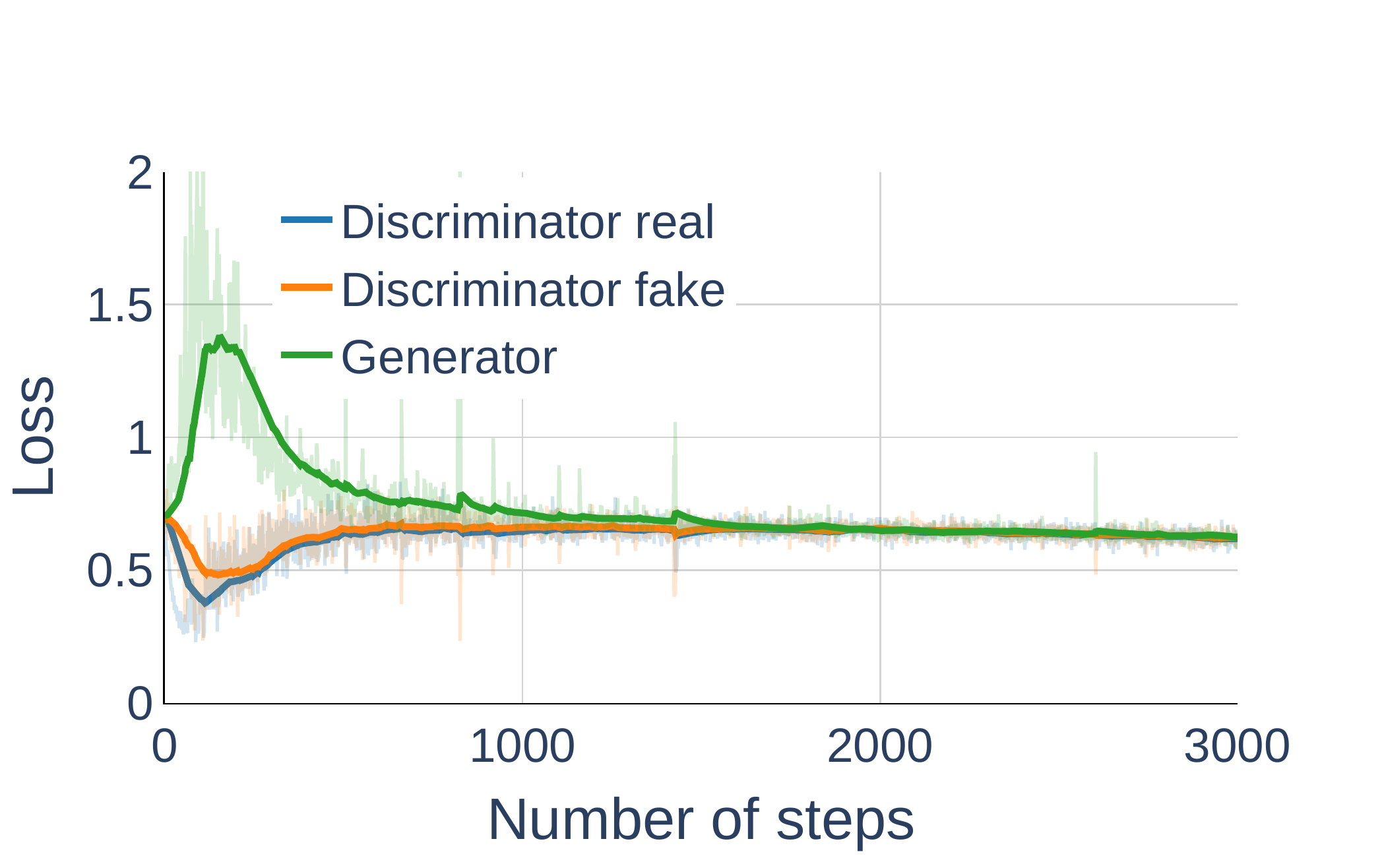}
    \centering
    \caption{Using improved training techniques}
\end{subfigure}
\caption{Training the GANCA structure with different settings. The first image shows the loss during training when using a simple GANCA structure, while the right image shows a loss graph when training with noise and label smoothing. A good loss for the generator and discriminator should converge between 0 and 1, similar to the right plot.}
\label{fig:ganca_comp}
\end{figure*}

As GANs are notoriously difficult to train,
several papers have been published to increase and stabilize the performance (\cite{arjovskyWassersteinGAN2017}, \cite{ImproveGANS_2}, \cite{gannoise2019}, \cite{dcganImproveGANs_3}, \cite{ImproveGANs_1}).
A good improvement to training, is to add noise to the input images of the discriminator and the output of the generator \cite{gannoise2019}, which makes it a lot harder to overfit for the discriminator. Adding even a small amount of noise leads to a drastically more stable performance. 
The second change we included, is to smooth the labels for training, 0 or 1 to a value similar to 0.1 or 0.9 as this tends to improve the performance of GAN models~\cite{ImproveGANs_1}.
A comparison using these training improvements can be found in Figure \ref{fig:ganca_comp}.

Additionally, using the WGAN loss helped to improve stability. 
The Wasserstein loss\cite{arjovskyWassersteinGAN2017} uses a critic network, which does not output probabilities but instead values between $[0,\infty)$.


\section{Results}

Without any major modifications, the standard NCA is able to fully reproduce the training images and keep them stable (persistence), after only 10k steps with a batch size of 16.

% ...

\begin{figure}
\centering
  \centering
  \includegraphics[width=.95\linewidth]{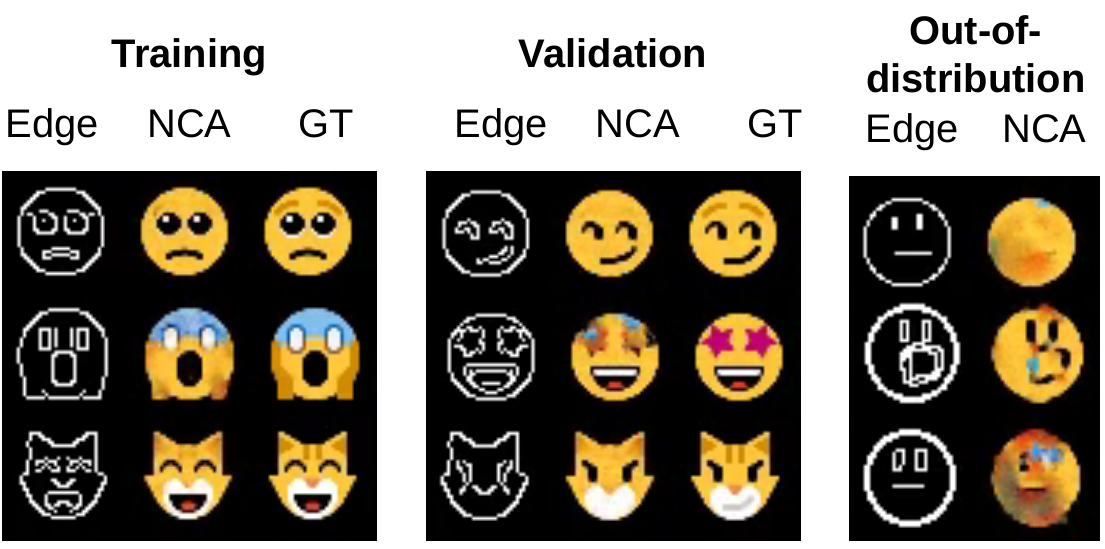}
  \caption{Output for training, standard validation and out-of-distribution data of the NCA after training for 10k steps on 50 different emoji faces. ``Edge'' is the edge image used as the input for the NCA, ``NCA'' is a single frame of the output of the NCA, while ``GT'' shows the target ground truth image.}
  \label{fig:nca_faces_50_output}
\end{figure}

These results are visualized in Figure \ref{fig:nca_faces_50_output}, where the NCA is trained on 50 different emoji faces with different edge images as input. The NCA is able to overfit on the training data, producing very good replicates to the ground truth images. Simple validation images also look good, while more complex images are missing color information. As the color information is never provided, this result is to be expected.
However, on out-of-distribution data, the NCA does not perform very well on any image. Many artifacts are introduced, and some details are removed.

Using the introduced method of the GANCA, the NCA can be trained adversarially to increase the out-of-distribution performance.

\begin{figure}
\centering
  \centering
  \vspace*{.85cm} 
  \includegraphics[width=.95\linewidth]{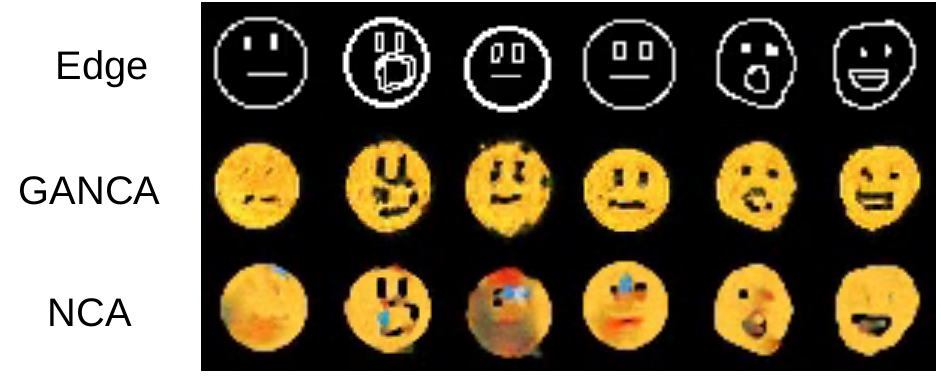}
  \vspace*{.10cm}
  \caption{Output of the GANCA and the standard NCA architecture after training on 50 different emojis. Edge images used as input here are out-of-distribution validation images. The GANCA shows significant improvement over the standard NCA training.}
  \label{fig:wgan_faces_output}
\end{figure}
As visualized in Figure \ref{fig:wgan_faces_output}, the performance on out-of-distribution data increases drastically in comparison to the standard NCA architecture. 
Additional images can be found in the appendix in Section \ref{sec:additional_img}.
Most images contain all the details and do not introduce any artifacts.

\section{Discussion}
The NCA showed very good results on generating different emojis with just a single NCA structure.
Thereby demonstrating that the architecture has enough capacity to reproduce all training images, to the extent of overfitting.
Even after overfitting, it was still able to produce decent-looking results on validation images.
Because the NCA is only able to use local information, the model did not fully overfit on the training data, and still performed decently on the validation images. 
This shows that the NCA is inherently able to generalize through the architecture alone. 

These generalization capabilities, specifically on out-of-distribution data, are further improved through unsupervised adversarial training. 

In addition, it should be noted that the number of parameters used in the generator architecture by using an NCA (around 20k parameters) is much smaller compared to other generators.
This is because a lot of information is placed in the environment and passed through the image step by step.
Because of this step by step information sharing, the NCA will need a longer time to produce the final output, compared to standard feed forward architectures.
However, \citet{sandlerImageSegmentationCellular2020} 
used asynchronous spatial updates for NCAs, allowing the use of a grid of processors without global synchronization, for a drastic decrease in computational time.

These experiments show a proof of concept, as training on high-quality images with big datasets, was not the main concern.
This is still possible and highly encouraged for future projects.
It is very promising to see such a drastic difference in results of the out-of-distribution data between the adversarial training and the standard supervised training.


\section{Conclusion}
We proposed two novel approaches for training NCAs, namely by training on multiple target images with a single model and by adversarial training with the GANCA architecture.
We demonstrated that a single NCA is capable of learning different emojis and is still able to perform well on validation images without additional color information.
Using existing training improvements from GAN architectures, NCAs can be trained in an adversarial fashion, which drastically improve performances on out-of-distribution data. 

\subsection{Future Work}
As NCAs have only recently been introduced and show great adaptability to a wide range of tasks, many possible directions are promising.
We strongly encourage further work that takes advantage of the fact that NCAs use an image as input and operate on the image step by step.

As a use case, NCA could be used for videos, e.g. for segmentation or generation, as each frame is connected to previous frames, which is inherently built in the architecture.

At last, working with user interactions is a promising field. 
By modifying the current state of the image, it allows for an easy way to interact with the model, as already showcased by \citet{mordvintsevGrowingNeuralCellular2020}. 
This type of interactive behavior is very intuitive with an NCA architecture, compared to a deconvolutional architecture in which the image is generated in a single forward pass. 
Moreover, extended by adversarial training, reacting to unpredictable behavior of users, could be a strong use case.

\bibliographystyle{plainnat}
\bibliography{refs}

\newpage
\appendix

\section{Additional Images}
\label{sec:additional_img}
\begin{figure}
\centering
  \centering
  \includegraphics[width=.95\linewidth]{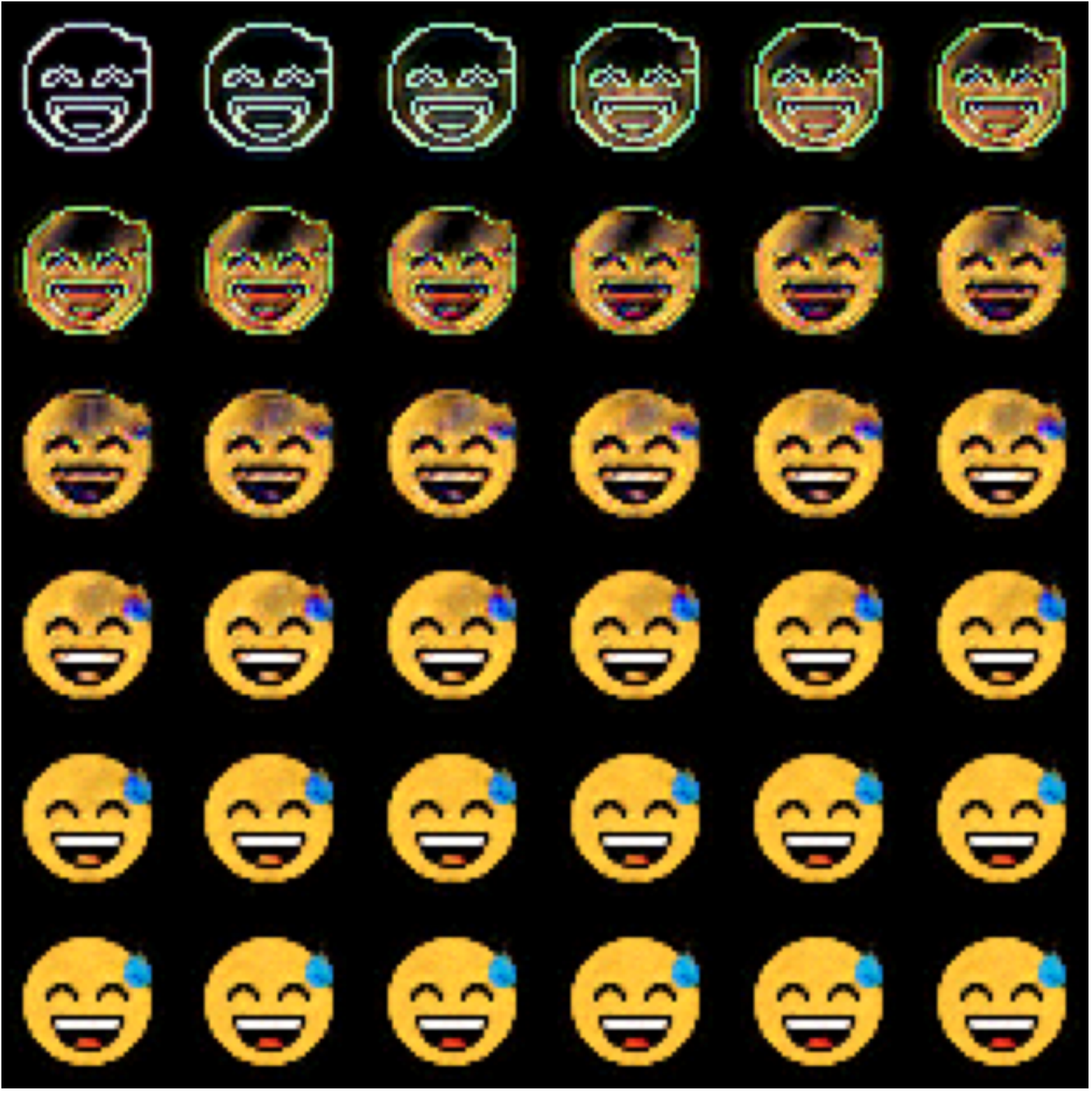}
  \caption{Output of an NCA frame by frame for a single emoji, showing 36 frames.}
  \label{fig:frame_by_frame}
\end{figure}

As including videos in pdfs is not possible, Figure \ref{fig:frame_by_frame} shows the output of every frame for 36 frames. It should be noted here that after around frame  24 the NCA has finished creating the emoji and will keep the image stable for the following frames. Additional frames could be added, but they do not show any or only minor changes. 

The following Figure \ref{fig:nature_out} shows the output of the NCA on a different dataset. 
The training images look very close to the ground truth images, as the NCA was able to overfit on the training data.
On the validation set, the performance decreases drastically. For example, the first image shows a tree, which the NCA mistakenly drew yellow instead of green. Because the NCA does not know the color of a tree, as no trees are present in the training set, this behavior is to be expected. 
Interestingly, the overall shape of each image stays consistent and only wrong colors are chosen.
This behavior is consistent to the training task of the NCA, as it needs to color some object in some specific way and keep it stable.

\begin{figure}
\centering
  \centering
  \includegraphics[width=.95\linewidth]{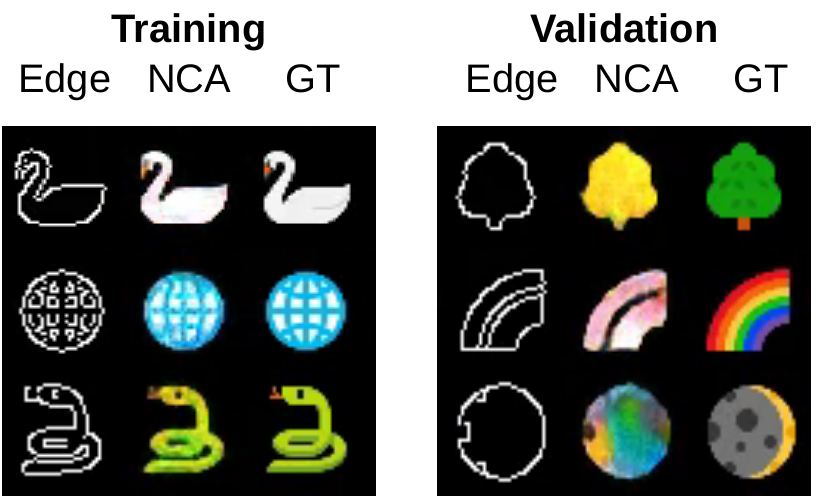}
  \caption{Output of the NCA after training on 50 different emojis from the ``nature'' category. The input (Edge) produces the output of the NCA (NCA) and is compared to the ground truth (GT).}
  \label{fig:nature_out}
\end{figure}

\begin{figure}
\centering
  \centering
  \includegraphics[width=.90\linewidth]{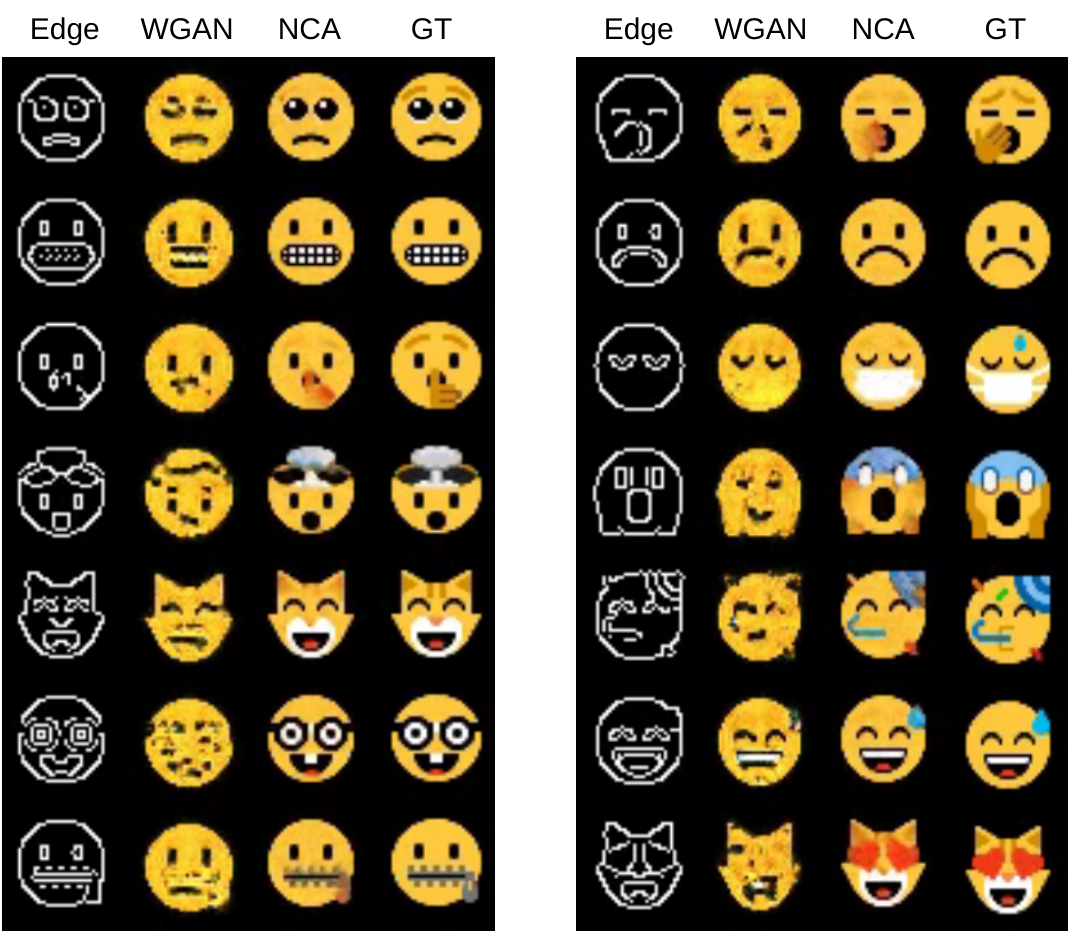}
  \caption{Additional outputs for the training dataset.}
  \label{fig:train_out_comp}
\end{figure}

\begin{figure}
\centering
  \centering
  \includegraphics[width=.90\linewidth]{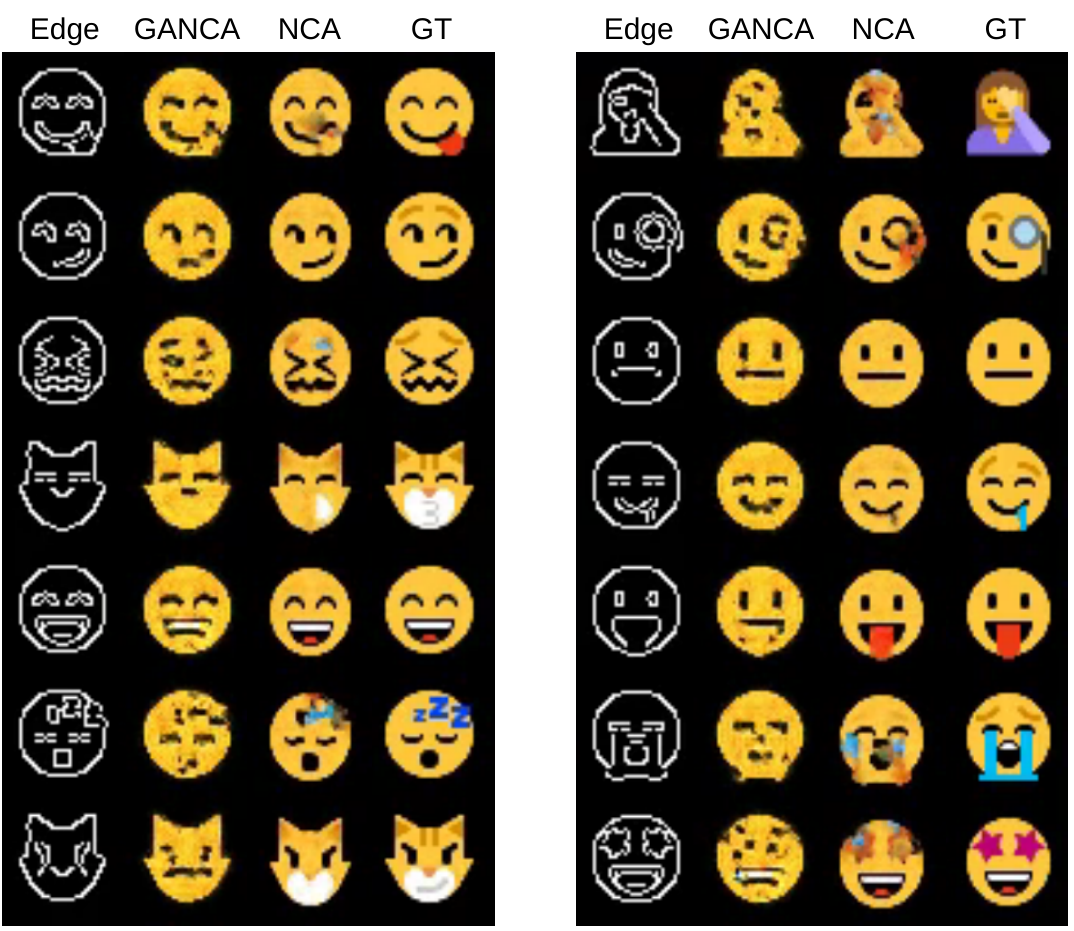}
  \caption{Additional outputs for the validation dataset}
  \label{fig:val_out_comp}
\end{figure}

\begin{figure}
\centering
  \centering
  \includegraphics[width=.90\linewidth]{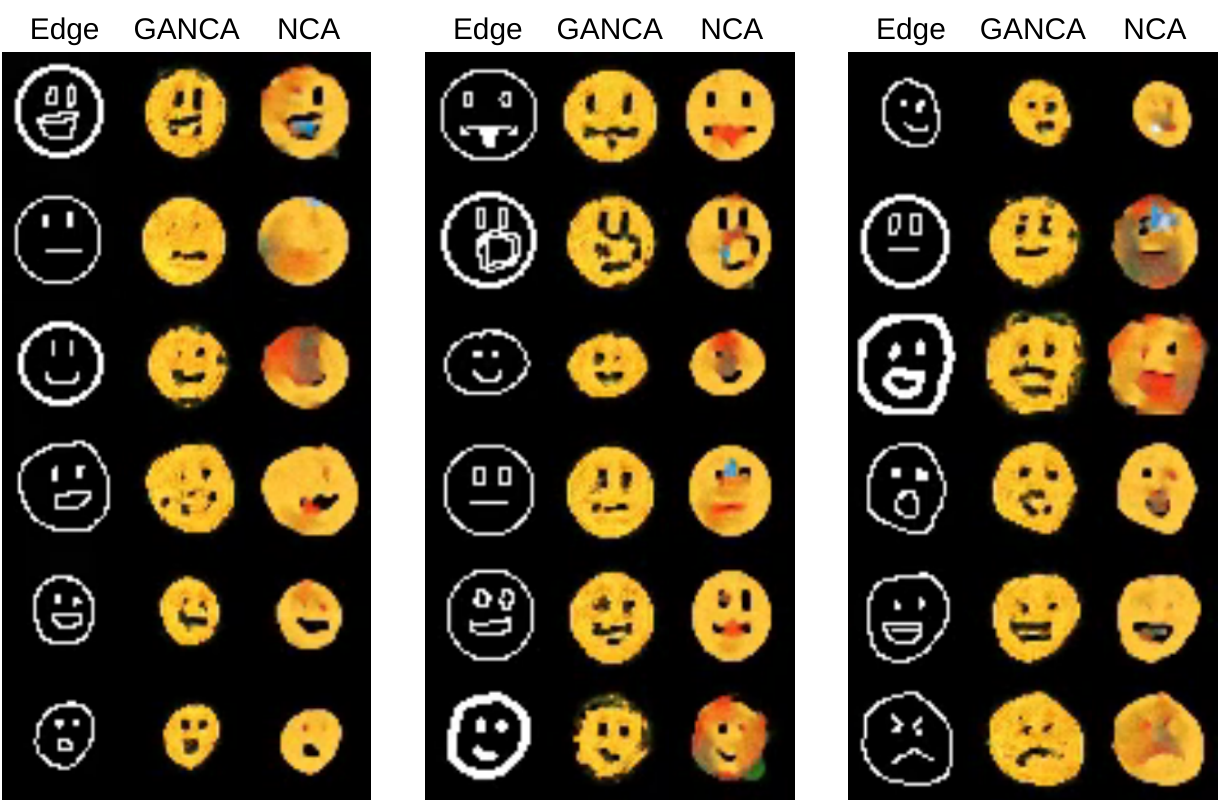}
  \caption{Additional outputs for the out-of-distribution dataset.}
  \label{fig:ood_out_comp}
\end{figure}

Additional outputs for training, validation and out-of-distribution images are provided in Figure \ref{fig:train_out_comp}, \ref{fig:val_out_comp} and \ref{fig:ood_out_comp}.
In each figure, the first column of images shows the input edge image, the GANCA output, the output of the NCA and the ground truth, if it exists. 
Similar to the example on ``nature'' images, the validation results for the supervised trained NCA are missing important color information, following the same reasoning.
Furthermore, the performance of the GANCA on the training and validation images can not directly be compared to the standard supervised NCA architecture. This is because through the unsupervised training, the GANCA was never trained to replicate the ground truth images.

\section{Additional Training Details}

\begin{figure}
\centering
  \centering
  \includegraphics[width=.95\linewidth]{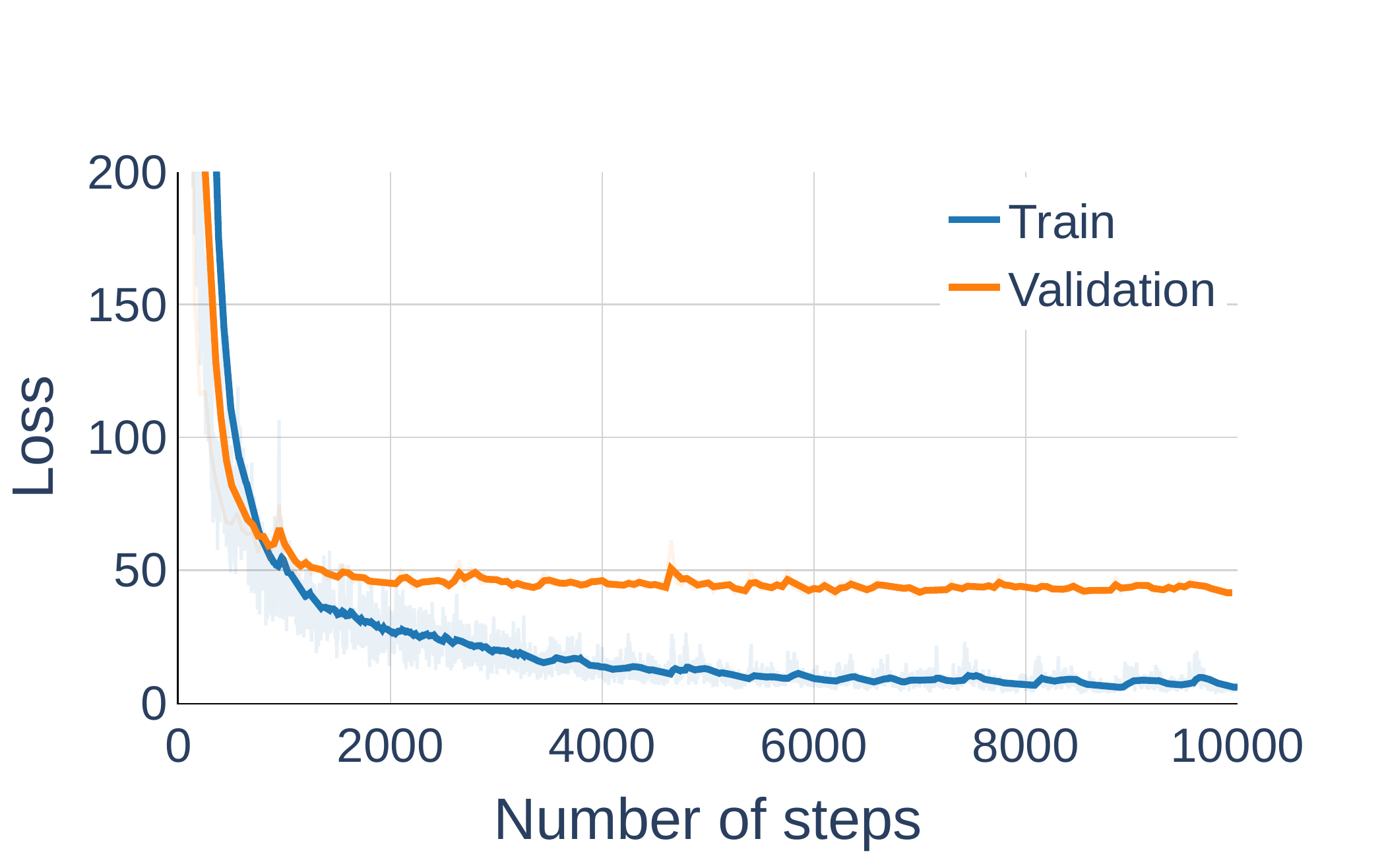}
  \caption{Training graph of the standard NCA learning to grow 50 different emoji faces. The loss used is the L2 loss.}
  \label{fig:nca_faces_loss}
\end{figure}

The additional graph in Figure \ref{fig:nca_faces_loss} shows the L2 loss when training an NCA supervised on 50 different emojis.
After around 1k steps with a batch size of 16, the NCA stops improving the validation loss. At this point, most images consist of a yellow mush, resulting in a decent average loss. 
The training loss keeps improving for the next steps, while the validation loss keeps mostly stable. 

In Figure \ref{fig:ganca_comp}, the training losses of two different GANCA architectures is compared. 
The improved GANCA uses the techniques introduced in 
Section \ref{sec:ganca}.

\end{document}